# Feature Extraction of ECG Signal Using HHT Algorithm

**Neha Soorma**
M.TECH (DC)
SSSIST, Sehore, M.P.,India

**Jaikaran Singh**
Department of Electronics and Communication,
SSSIST, Sehore, M.P. India

**Mukesh Tiwari**
Department of Electronics and Communication,
SSSIST, Sehore, M.P. India

*Abstract*— **This paper describe the features extraction algorithm for electrocardiogram (ECG) signal using Huang Hilbert Transform and Wavelet Transform. ECG signal for an individual human being is different due to unique heart structure. The purpose of feature extraction of ECG signal would allow successful abnormality detection and efficient prognosis due to heart disorder. Some major important features will be extracted from ECG signals such as amplitude, duration, pre-gradient, post-gradient and so on. Therefore, we need a strong mathematical model to extract such useful parameter. Here an adaptive mathematical analysis model is Hilbert-Huang transform (HHT). This new approach, the Hilbert-Huang transform, is implemented to analyze the non-linear and non-stationary data. It is unique and different from the existing methods of data analysis and does not require an a priori functional basis. The effectiveness of the proposed scheme is verified through the simulation.**

## I. Introduction

Heart structure is a unique system used to generate an ECG signal independently via heart contraction. The needs of technology and computerized analysis usage has exhorted researchers, professionals, engineers and other expert people combining their efforts together in implementing quality diagnosis tools. The term quality has been interpreted as easier and faster analysis, lack maintenance, high efficient as well as low in the cost. To analyze ECG signals focusing on real peaks recognition since it provides valuable information to doctors regarding heart diagnosis [1]. With the help of ECG, the electrical activity within the heart can be easily detected from the outside of the body. When the ECG is abnormal it is called Arrhythmia. The patterns of the waveform change due to abnormalities of the heart [2]. Most of the clinically useful information in the ECG is found in the intervals and amplitudes defined by its features (characteristic wave peaks and time durations). The development of accurate and quick methods for automatic ECG feature extraction is of major importance, especially for the analysis of long recordings (Holters and ambulatory systems). In fact, beat detection is necessary to determine the heart rate, and several related arrhythmias such as Tachycardia, Bradycardia and Heart Rate Variation; it is also necessary for further processing of the signal in order to detect abnormal beats [3]. Producing algorithms for the automatic extraction of the ECG features. is complicated due to the time-varying nature of the signal resulting of variable physiological conditions and the presence of noise.

In recent years, the wavelet transform emerged in the field of image/signal processing as an alternative to the well-known Fourier Transform (FT) and its related transforms, namely, the Discrete Cosine Transform (DCT) and the Discrete Sine Transform (DST). In the Fourier theory, a signal (an image is considered as a finite 2-D signal) is expressed as a sum, theoretically infinite, of sines and cosines, making the FT suitable for infinite and periodic signal analysis. For several years, the FT dominated the field of signal processing, however, if it succeeded well in providing the frequency information contained in the analysed signal; it failed to give any information about the occurrence time. This shortcoming, but not the only one, motivated the scientists to scrutinise the transform horizon for a “messiah” transform. The first step in this long research journey was to cut the signal of interest in several parts and then to analyse each part separately. The idea at a first glance seemed to be very promising since it allowed the extraction of time information and the localisation of different frequency components. This approach is known as the Short-Time Fourier Transform (STFT). The fundamental question, which arises here, is how to cut the signal? The best solution to this dilemma was of course to find a fully scalable modulated window in which no signal cutting is needed anymore. This goal was achieved successfully by the use of the wavelet transform.

However a new approach, the Hilbert-Huang transform, is also developed to analyse the non-linear and non-stationary data. It is unique and different from the existing methods of data analysis and does not require an a priori functional basis. By using HHT method, things will be much simpler, time-savings as well as reducing the needs of human efforts as machine has been trained to perform the desired workload.

Because of the distinct characteristics of HHT, it has attracted considerable research interest in exploring its potential as a frequency identification tool. A straightforward method could be that, after application of HHT to a signal, comparisons are made between Fourier spectra of the obtained IMFs and that of the original signal to find out the relationships between IMFs and vibration modes. Then by computing the amplitude weighted average frequencies based on the Hilbert spectra, modal frequencies can be identified. Besides, Yang et al. proposed a method in which, before they are analyzed by HHT, the signals are processed by some pre-selected band pass filters, the thresholds of which are determined by referring to the Fourier spectra of the signals.

This paper describes the features extraction algorithm for electrocardiogram (ECG) signal using wavelet transform as well as HHT. The purpose of feature extraction of ECG signal would allow successful abnormality detection and efficient prognosis due to heart disorder. Therefore, we need a strong mathematic model to extract such useful parameter.

This paper is organized as follows: Section II deals with proposed methodology to extract features of any ECG signal. Hilbert Huang Transform algorithm is explained in section III. Effectiveness of the HHT algorithm is illustrated in section IV to extract features of ECG signal. Section V concludes the paper.

## I. Wavelet Transform

A Wave is an oscillating function of time or space, Wavelets are localized waves and they have their energy concentrated in time or space. The Transform of a signal is another form of representing the signal. It does not change the information content present in the signal. The Wavelet Transform provides a time- frequency representation of the signal and is well suited to the analysis of non-stationary signals [9] such as ECG. A Wavelet Transformation uses multi resolution technique by which different frequencies are analysed with different resolutions. A Wavelet Transform, at high frequencies, gives good time resolution and poor frequency resolution, while at low frequencies the Wavelet Transform gives good frequency resolution and poor time resolutions.

One of the most frequently and commonly used Wavelet Transformation is the Discrete Wavelet Transformation (DWT). Mathematically a Discrete Wavelet Transform can be represented as

Where x(t) is the signal to be analyzed. Ψ(t) is the mother wavelet or the basis function. All the wavelet functions used in the transform are derived from the mother wavelet through wavelet translation (n) and scaling (m). The translation parameter n refers to the location of the wavelet function as it is shifted through the signal and the scale parameter m corresponds to frequency information. Large scales (low frequencies) dilate the signal and provide detailed information hidden in the signal, while small scales (high frequencies) compress the signal and provide global information about the signal. The wavelet transform merely performs the convolution operation of the signal and the basis function.

## II. Hilbert- Huang Transform

To analyse the data which is nonlinear and non-stationary, various attempts such as Spectrograms, Wavelet analysis, and the neural network etc have been made, but the Hilbert- Huang Transform approach is unique and different from the existing methods. The fundamental parts of the HHT are the Empirical Mode Decomposition (EMD) and Hilbert spectral analysis method. By EMD method, any complicated problem related to engineering, biomedical, financial and geophysical data can be resolve due to an adaptive time-frequency analysis. In this process data set can be decomposed into a finite and often small number of components, which is a collection of intrinsic mode functions (IMF). The Hilbert spectral analysis (HSA) provides a method for examining the IMF's instantaneous frequency data as functions of time that give sharp identifications of embedded structures [5].

### *Empirical Mode Decomposition (EMD) Algorithm*

The EMD method is well suited for analysing time-series data representing non stationary and nonlinear processes. This method could decompose any time-varying data into a finite set of functions called "intrinsic mode functions" (IMFs)[6]. An IMF can have variable amplitude and frequency along the time axis. The procedure of extracting an IMF is called sifting. An IMF is a function that satisfies the following requirements:

1. In the whole data set, the number of extrema and the number of zero-crossings must either be equal or differ at most by one.

2. At any point, the mean value of the envelope by the local maxima and the envelope defined by the local minima is zero.

The sifting process is as follows:

(1) Identify the extrema (both maxima and minima) of data $x(t)$;
(2) Generate the upper and lower envelopes $h(t)$ and $l(t)$, respectively, by connecting the maxima and minima points separately with cubic spline interpolation;
(3) Determine the local mean $m_1(t) = (h(t)+l(t))/2$;
(4) IMF should have zero local mean, subtract out $m_1$ from $x(t)$,

$$h_1(t) = x(t) - m_1(t) \qquad (1.1)$$

(5) Test whether $h_1(t)$ is an IMF or not;

$$h_2(t) = h_1(t) - m_2(t) \qquad (1.2)$$

(6) Repeat steps 1 to 5 and end up with an IMF $h_1(t)$.

Once the first IMF is derived, define $C_1(t) = h_1(t)$, this is the finest temporal scale in the time-series data, that is, the shortest period component of the data $x(t)$. To find all the IMFs, generate the residue $r_1(t)$ of the data by subtracting out $C_1(t)$ from the data as

$$r_1(t) = x(t) - C_1(t) \qquad (1.3)$$

The residue now contains information about the components for longer period; it is treated as the new data and is resifted to find additional components. The sifting process will be continued until it meets a stopping criterion yielding the subsequent IMFs as well as residues and the result is

$$r_1(t) - C_2(t) = r_2(t),$$
$$r_2(t) - C_3(t) = r_3(t),$$
$$r_{n-1}(t) - C_n(t) = r_n(t) \qquad (1.4)$$

where $r_n(t)$ becomes a constant, a monotonic function, or a function with only maxima and one minima from which no more IMF can be derived. Two different criteria have been used: The first one was used in Huang et al. This stoppage criterion is determined by using a Cauchy type of convergence test. Specifically, the test requires the normalized squared difference between two successive sifting operations defined as

$$SD_k = \frac{\sum_{t=0}^{T} |h_{k-1}(t) - h_k(t)|^2}{\sum_{t=0}^{T} h^2_{k-1}(t)} \qquad (1.5)$$

to be small. If this squared difference $SD_k$ is smaller than a predetermined value, the sifting process will be stopped. Second criterion based on the agreement of the number of zero-crossings and extrema. Specifically, a S-number is pre-selected. The sifting process will stop only after S consecutive times, when the numbers of zero-crossings and extrema stay the same and are equal or differ at most by one. At the end of the decomposition the signal $x(t)$ is represented as

$$x(t) = \sum_{i=1}^{n} C_i(t) + r_n(t) \qquad (1.6)$$

where $n$ is the number of IMFs and $r_n(t)$ is the final residue.
Once the signal is decomposed to a series of IMFs and a residue, HSA is applied to each IMF. In fact, for any of the IMFs $C_i(t)$, the corresponding $\check{C}_i(t)$ is computed by Hilbert transform. An analytic signal $Z_i$ (t) is then formed, the magnitude of which is the instantaneous magnitude $A_i(t)$, and the derivative of the phase angle of which, $\theta_i(t)$, is the instantaneous frequency $\omega_i(t)$. Note that to avoid meaningless negative frequencies, the phase angle $\theta_i(t)$ must be unwrapped before the derivative is taken.
Procedures of HSA are shown in Eqs.( 1.7)-(1.10):

$$Z_i(t) = C_i(t) + i\,\check{C}_i(t) \qquad (1.7)$$

$$A_i(t) = |Z_i(t)| \qquad (1.8)$$

$$\tan \theta_i(t) = \frac{\check{C}_i(t)}{C_i(t)} \qquad (1.9)$$

$$\omega_i(t) = \frac{d\,\theta_i(t)}{dt} \qquad (1.10)$$

where the subscript i = 1, 2,..........................., n.
Upon finishing EMD and HSA, and by dropping the residue $r_n(t)$, the original signal x(t) can now be expressed as

$$x(t) = \mathrm{Re}\left[\sum_{i=1}^{n} A_i(t) \exp\left(i\int \omega_i(t)\,dt\right)\right] \qquad (1.11)$$

By comparing Eq. (2.10) with the Fourier series representation

$$x(t) = \sum_{i=-\infty}^{\infty} C_i(t)\, e^{j\omega_i(t)} \qquad (1.12)$$

where $C_i(t)$ and $\omega_i(t)$ are time-independent constants for a given x(t), it is clear that HHT is characterized by expressing a given signal by the sum of a finite number of adaptive base functions.

## III. Simulation Results

The ECG signal is simulated and used as a data or test signal. Two types of ECG signals (normal and abnormal) have been simulated and analyzed by wavelet transform as well as by HHT by extracting their features. Haar wavelet is used as a basis function to identify the frequency components of the simulated ECG signals as shown in figures 2 and 10. HHT has also been implemented for the features extraction of the normal and abnormal ECG using empirical mode decomposition. The IMF's generated after every iteration are shown in figures 3-8 for normal ECG and in figures 11-17 for abnormal ECG. All the generated IMF's are represented along with their frequency spectrum. We get signal with lower frequency content after successive stages of sifting of the signal which have high-frequency content.

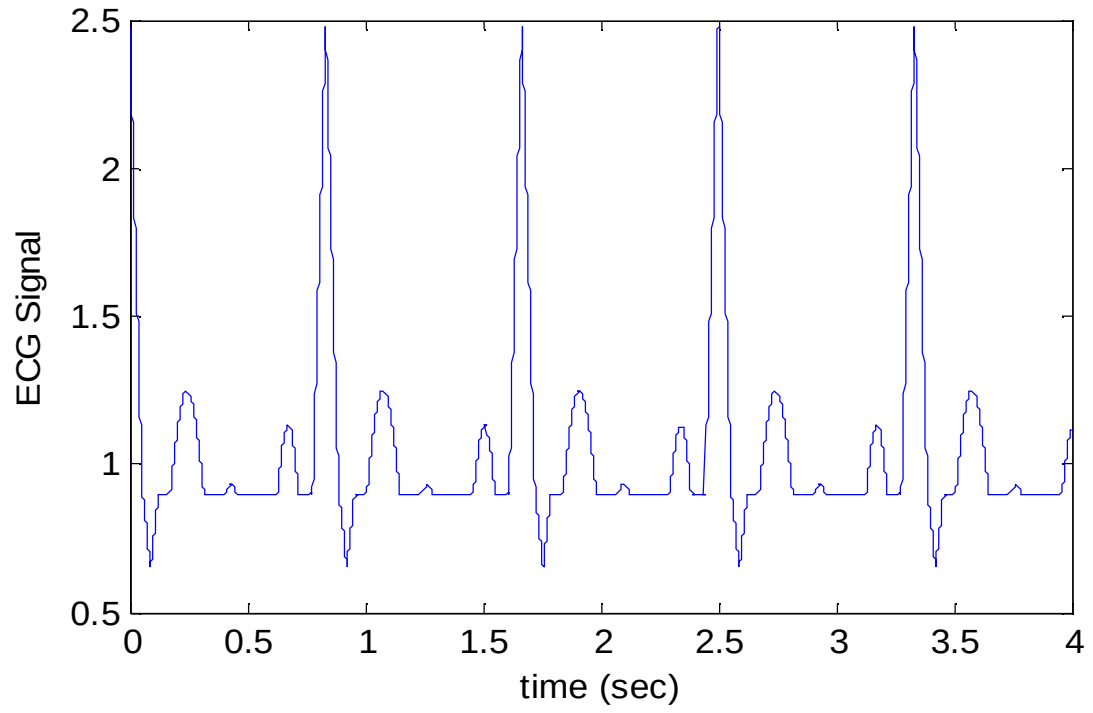

Fig 1. ECG signal of a normal patient

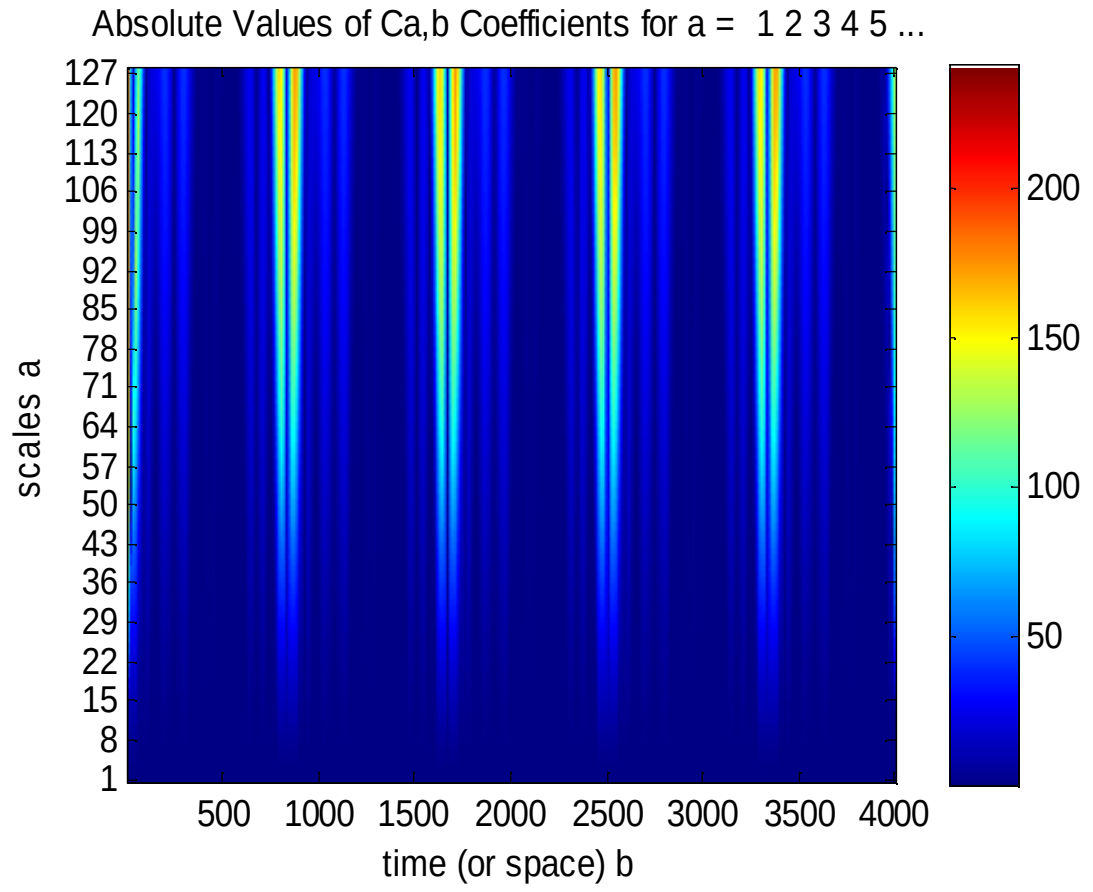

Fig 2. Wavelet transform of ECG Signal

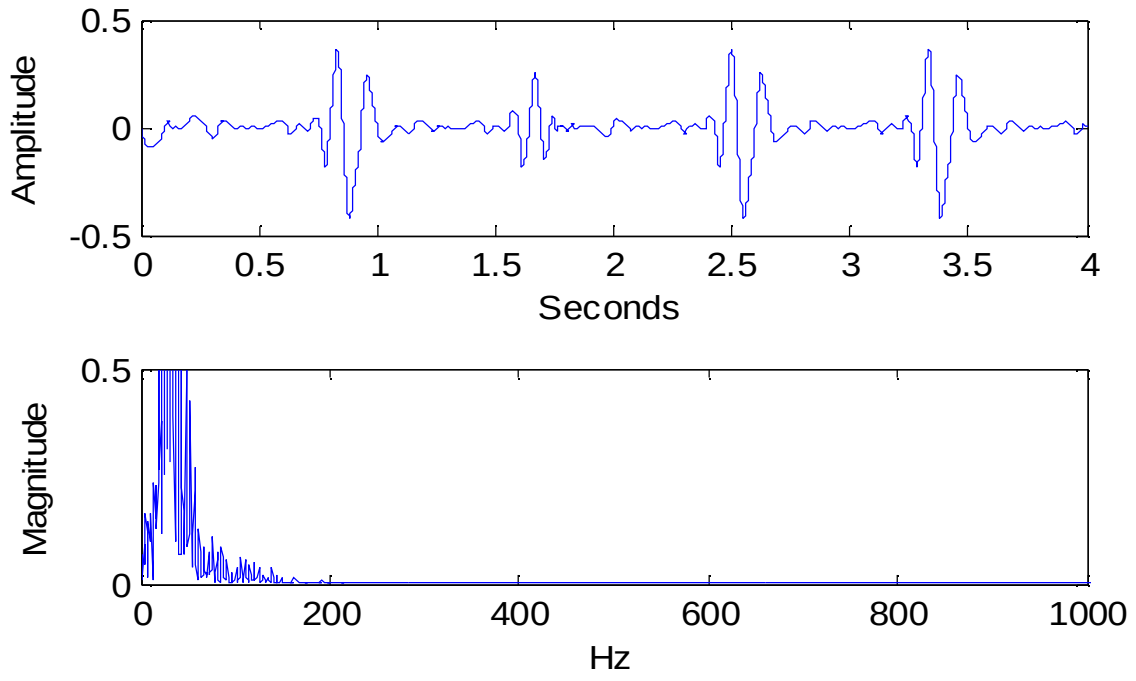

Fig 3. First IMF and its spectrum

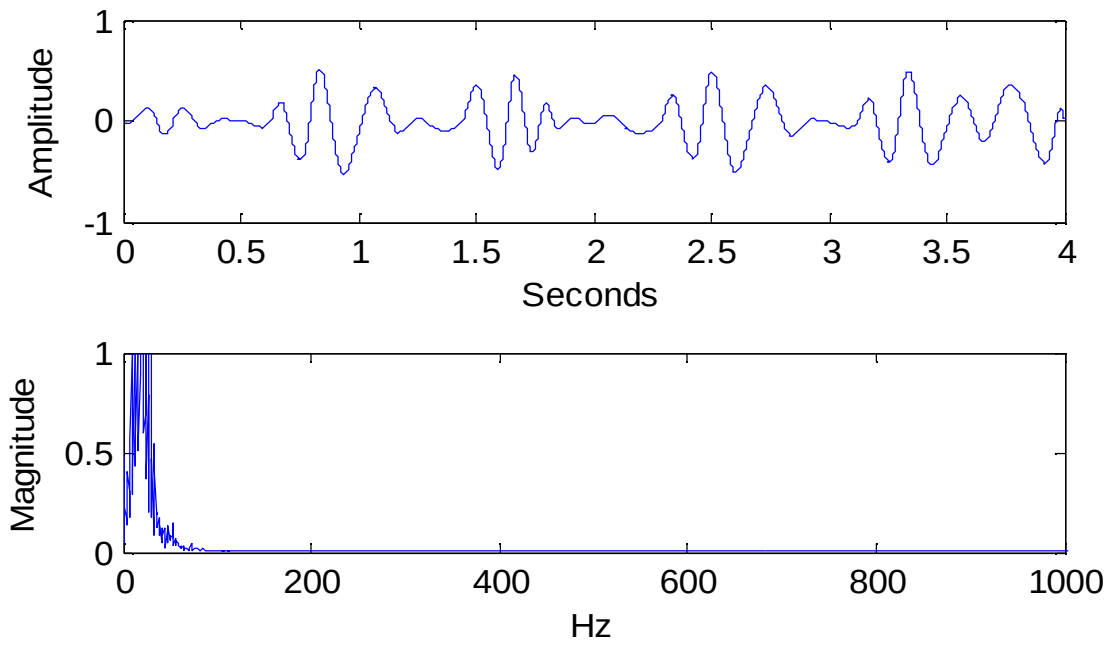

Fig 4. Second IMF and its spectrum

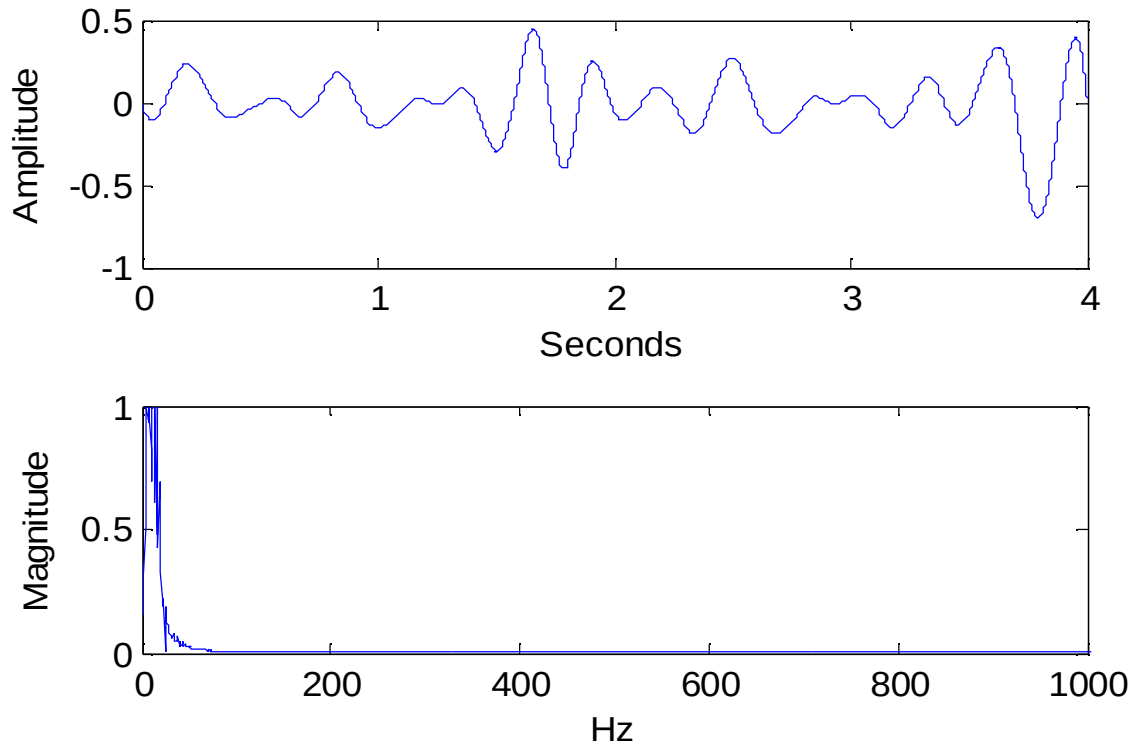

Fig 5. Third IMF and its spectrum

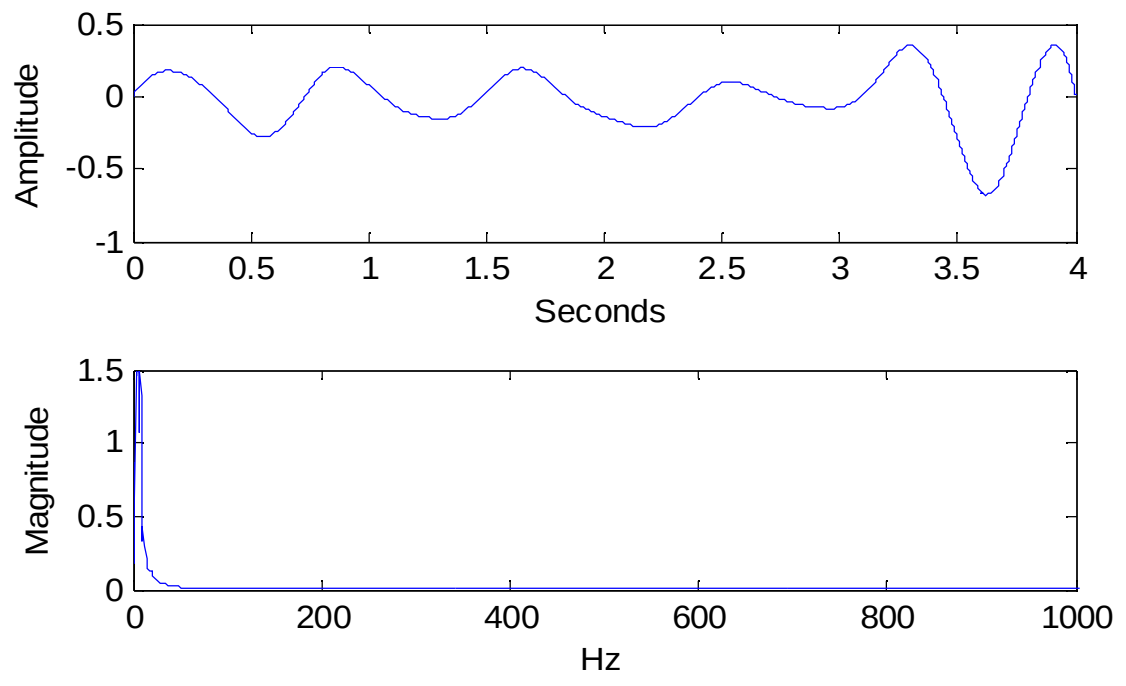

Fig 6. Fourth IMF and its spectrum

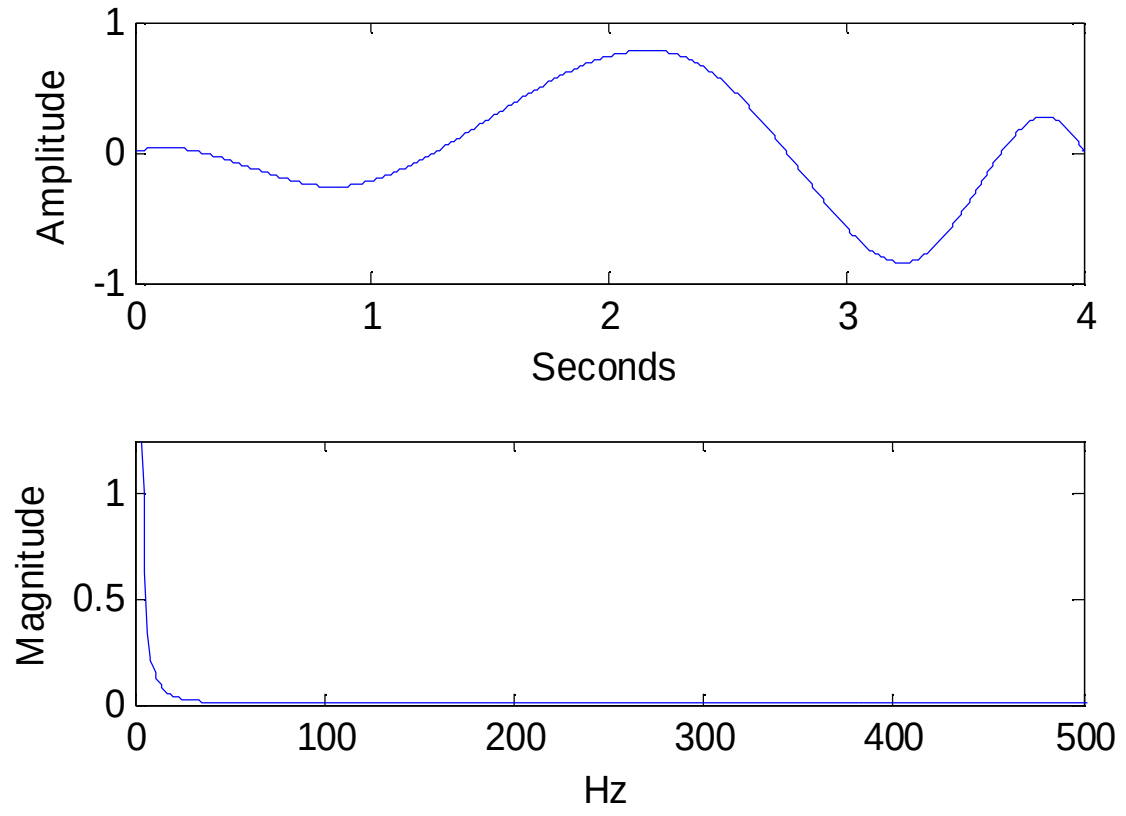

Fig 7. Fifth IMF and its spectrum

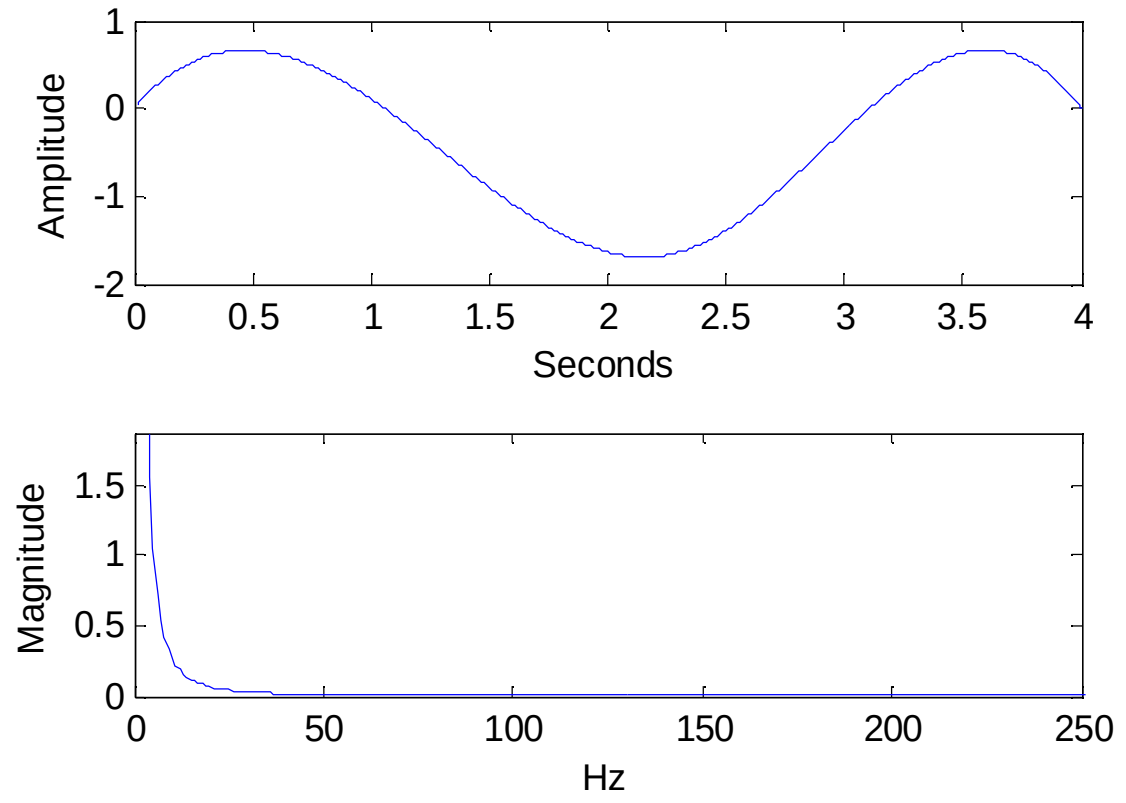

Fig 8. Last monotone IMF and its spectrum

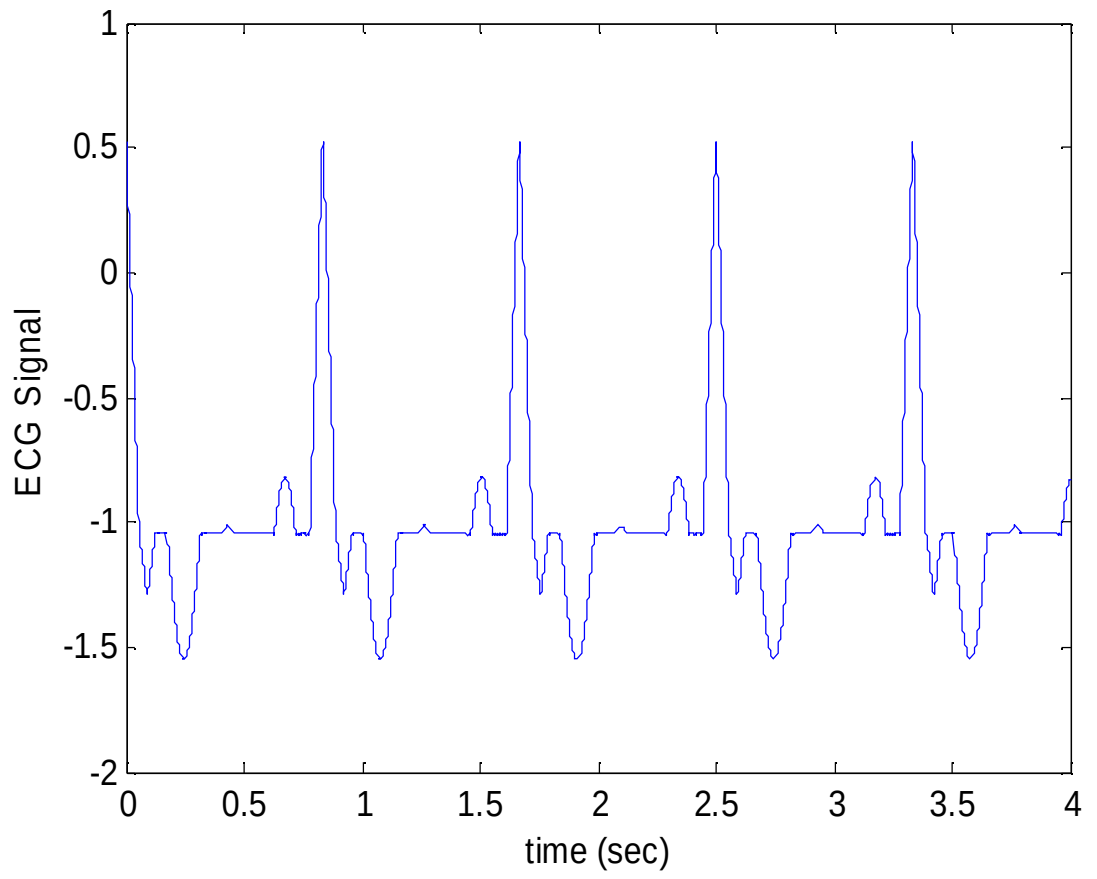

Fig 9. ECG Signal of abnormal patient

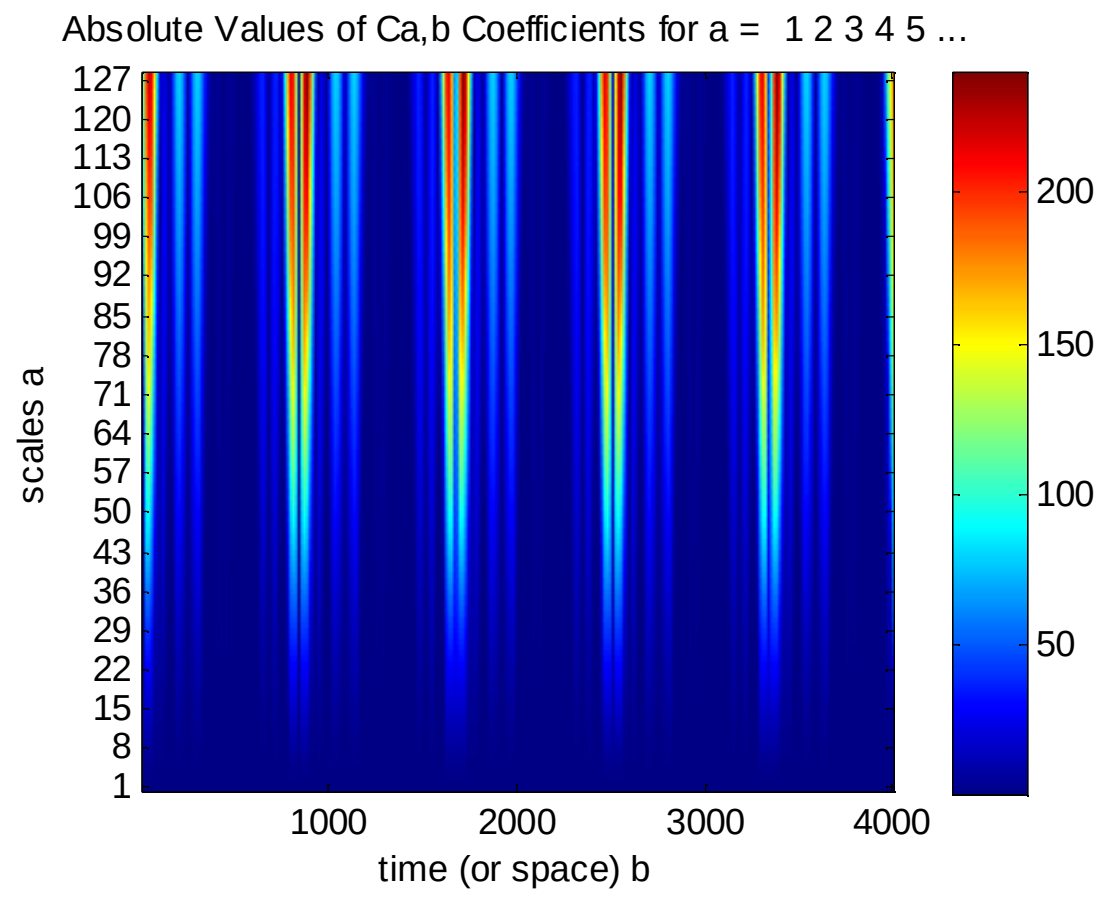

Fig 10. Wavelet Transform of abnormal ECG signal

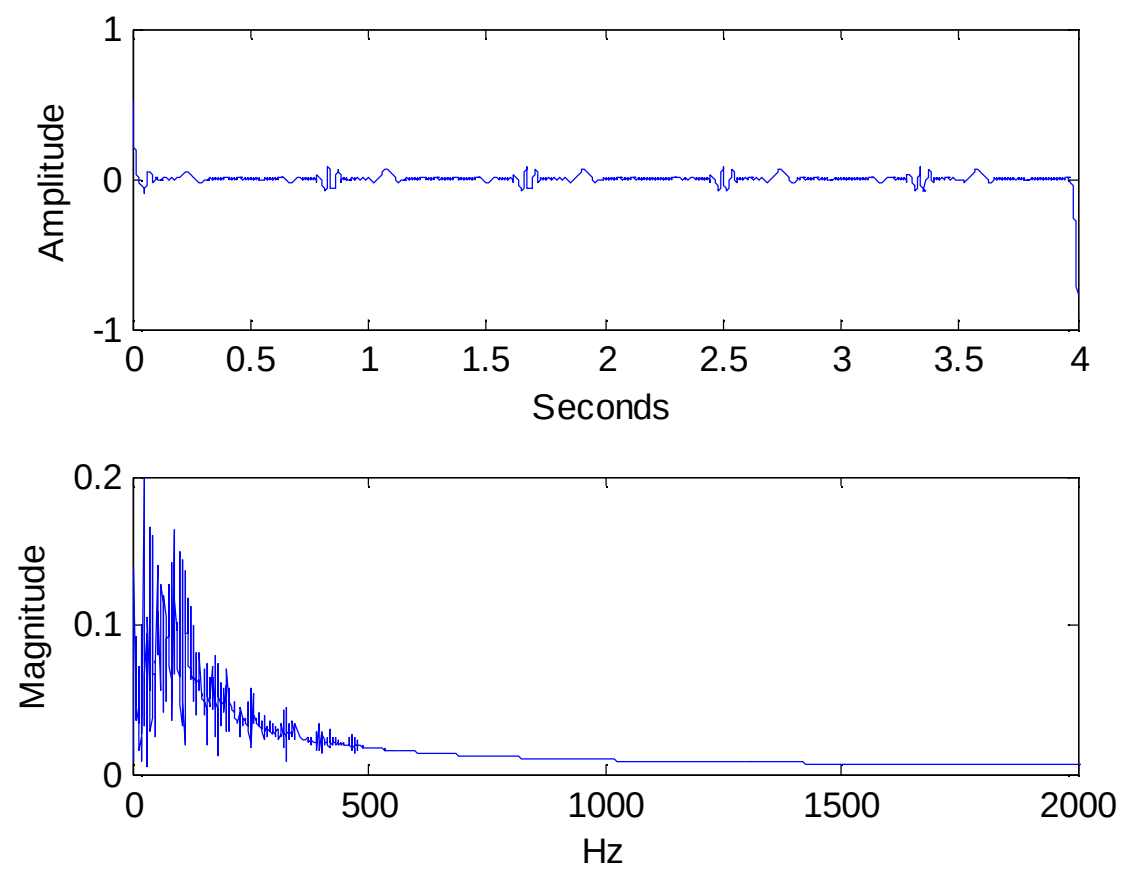

Fig 11. First IMF and its spectrum

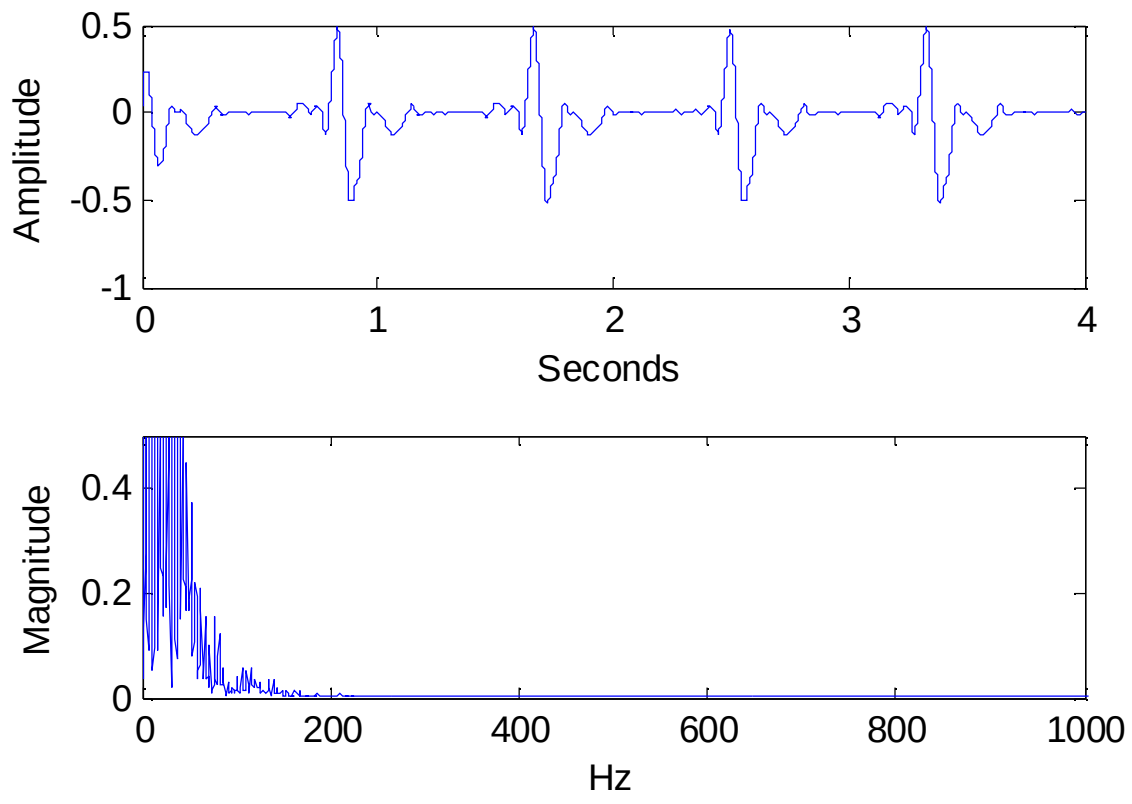

Fig 12. Second IMF and its spectrum

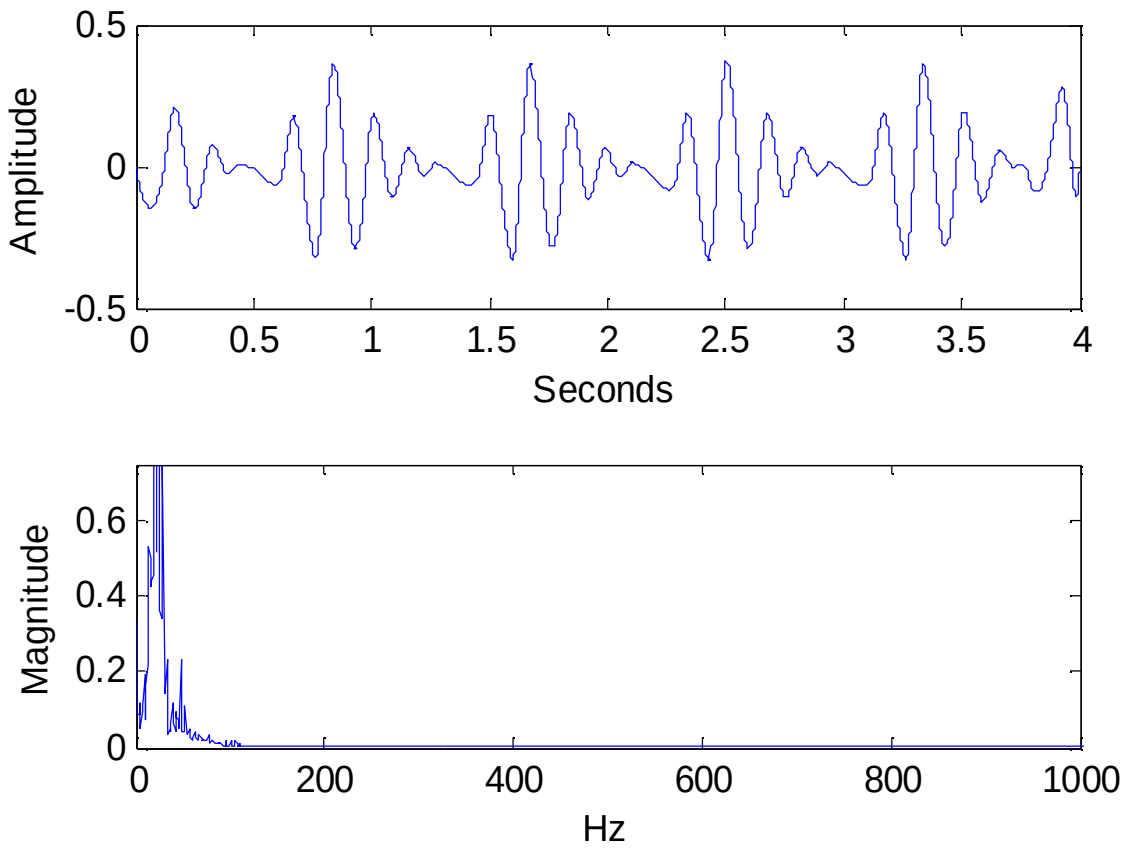

Fig 13. Third IMF and its spectrum

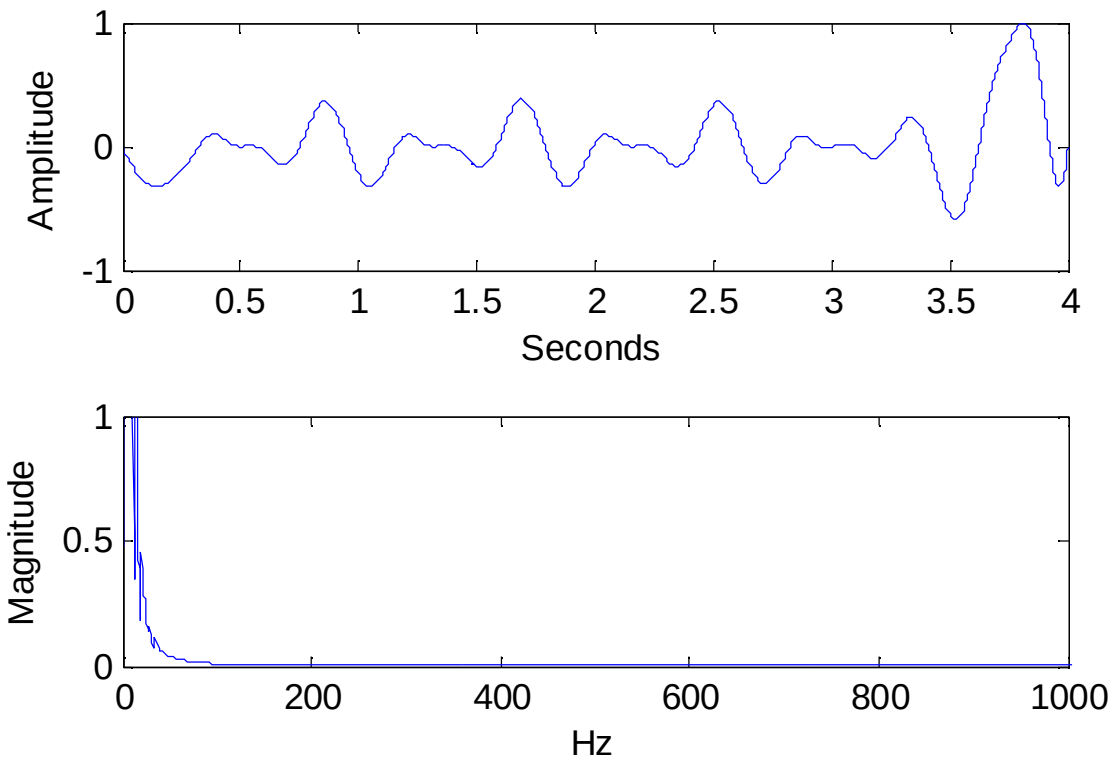

Fig 14. Fourth IMF and its spectrum

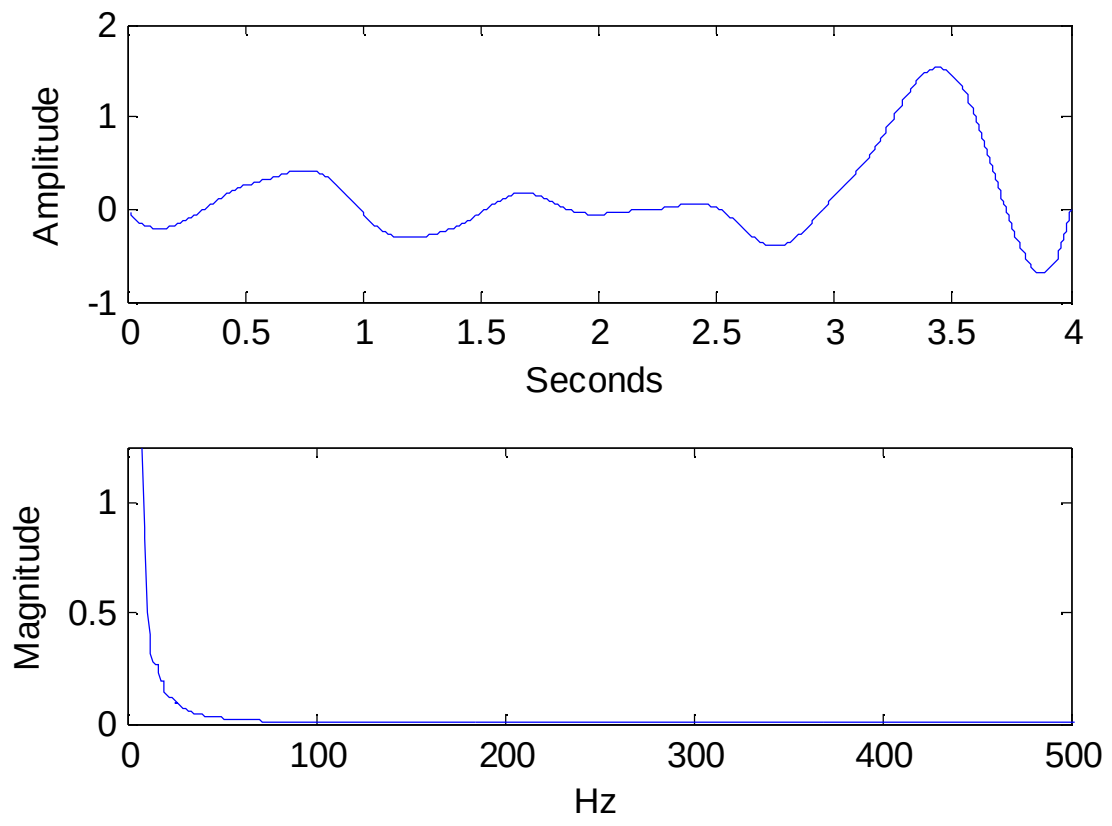

Fig 15. Fifth IMF and its spectrum

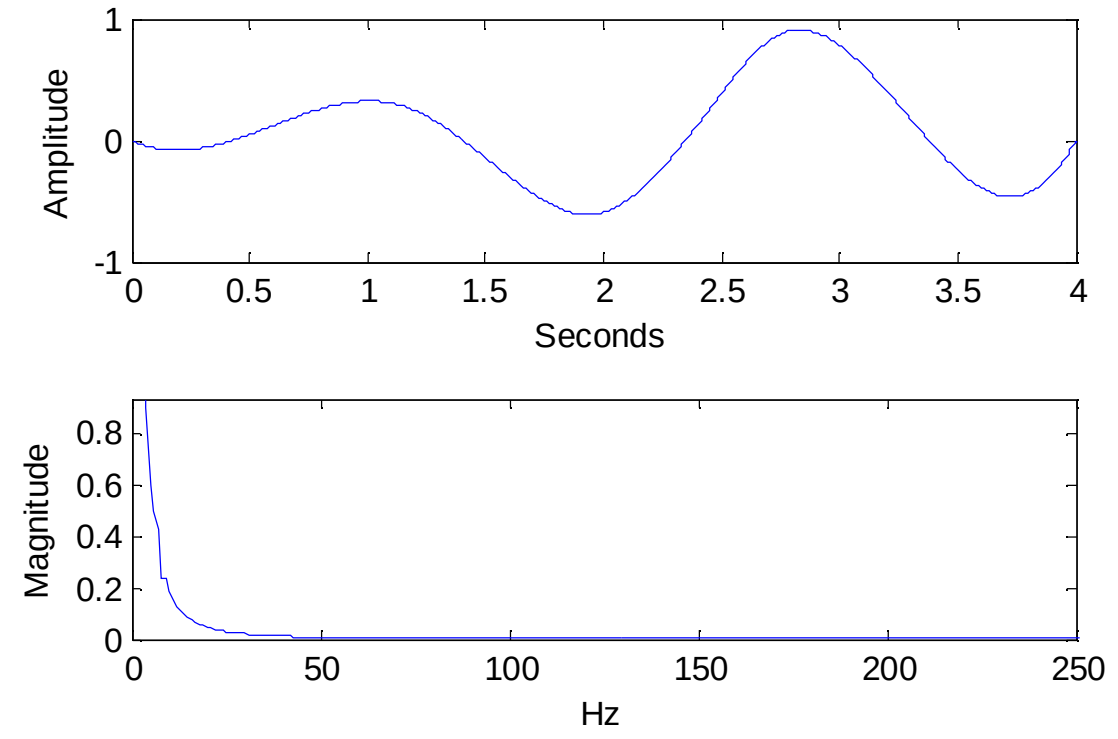

Fig 16. Sixth IMF and its spectrum

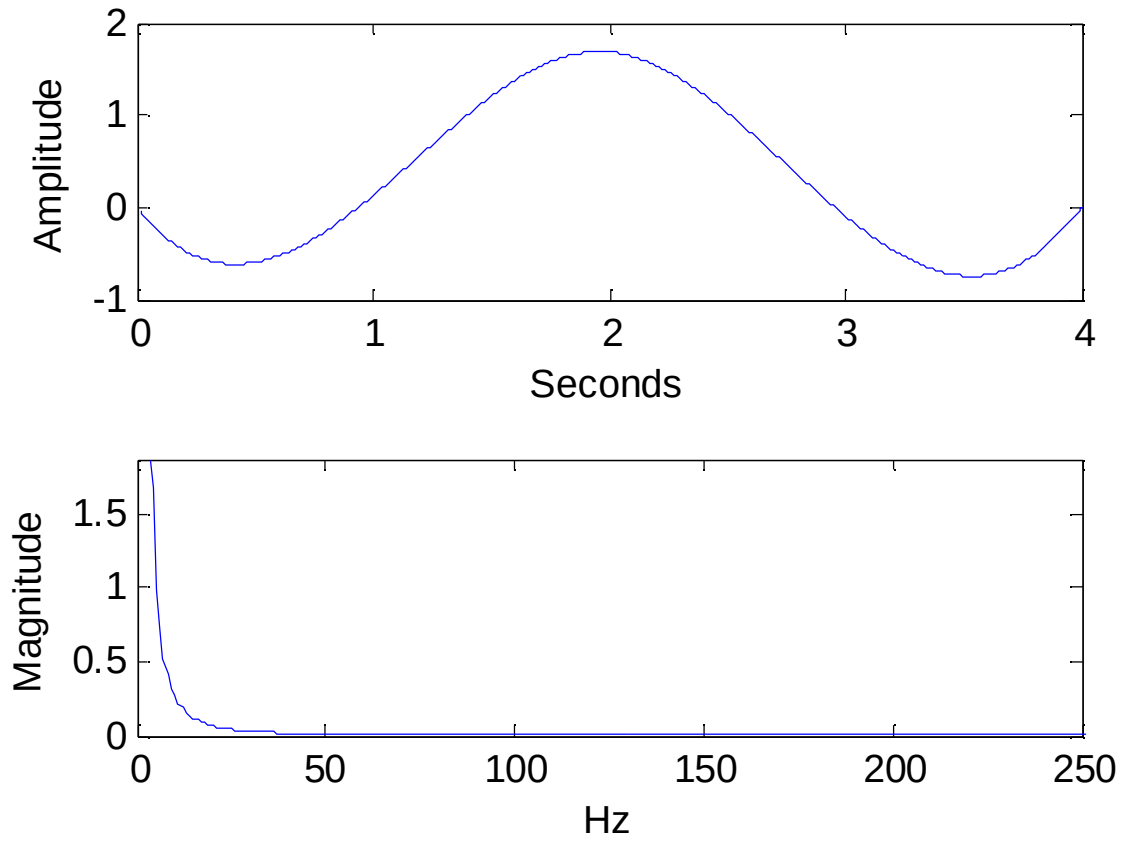

Fig 17. Last monotone IMF and its spectrum

## IV. Conclusion

Huang Hilbert Transform and wavelet transform have been implemented in this work to extract the features of ECG signal (normal and abnormal). The HHT algorithm is the best suited algorithm for extracting features of any signal. The lower frequency content of any signal is generated using HHT algorithm which is useful to analyze any nonlinear signal. When down sampling of a nonlinear signal is required then this HHT algorithm is useful. Non-uniform sampling techniques may be useful here, although they appear to require more complex up-sampling procedures to restore their original sampling rates than do uniformly sampled signals. Wavelet transform is requiring a fixed basis function for the analysis and found to be complex from the point of view of implementation. Simulation results are reflecting the efficiency of the HHT and wavelet transform technique in features extraction of ECG signal.

## References

[1] Guodong Tang and Aina Qin, "ECG Denoising based on Empirical Mode Decomposition," 9th International Conference for Young Computer Scientists, pp. 903-906, 2008.

[2] P. Zarychta , F.E. Smith, S.T. King, A.J.Haigh ,A. Klinge, S. Stevens , J. Allen, "Body surface potential mapping for detection of myocardial infarct sites," in Proc. IEEE comput. Cardiol, pp.181-184, 2007.

[3] Osowski S, Linh TH, "ECG beat recognition using fuzzy hybrid neural network", IEEE Trans Biom Eng, Vol; 48, pp: 1265-1271, 2001.

[4] P. Chazal, M. O'Dwyer, R.B. Reilly, "Automatic classification of heartbeats using ECG morphology and heartbeat interval feature", IEEE Trans Biom Eng, Vol; 51, pp: 1196-1206, 2004.

[5] M. Kania, M Fereniee, R. Maniewski, "Wavelet Denoising for multi-lead high resolution ECG signal", Measurement Science Review, Vol: 7, No: 2, No.4, 2007.

[6] S. Karpagachelvi, M. Arthanari, M. Sivakumar, "Classification of ECG signals using extreme Learning Machine", Computer and Information Science, Canadian Centre of Science and Education, Vol.4, No. 1; 2011.

[7] N. E. Huang, Z. Shen, S. R. Long, M. C. Wu, H. H. Shih, Q. Zheng, N.-C. Yen, C. C. Tung, and H. H. Liu, "The empirical mode decomposition and the Hilbert spectrum for nonlinear and nonstationary time series analysis", Proc. Roy. Soc. Lond, vol. A,454,. 903–995, 1998.

[8] Jerritta, S. ; Murugappan, M. ; Wan, K. ; Yaacob, S., "Emotion recognition from electrocardiogram signals using Hilbert Huang Transform" IEEE Conference on Sustainable Utilization and Development in Engineering and Technology, pp. 82 – 86, 2012.

[9] S. C. Saxena, A. Sharma, and S. C. Chaudhary, ―Data compression and feature extraction of ECG signals,‖ International Journal of Systems Science, vol. 28, no. 5, pp. 483-498, 1997.

[10] B. Castro, D. Kogan, and A. B. Geva, ―ECG feature extraction using optimal mother wavelet, 21st IEEE Convention of the Electrical and Electronic Engineers in Israel, pp. 346-350, 2000.

[11] C. Alexakis, H. O. Nyongesa, R. Saatchi, N. D. Harris, C. Davies, C. Emery, R. H. Ireland, and S. R. Heller, ―Feature Extraction and Classification of Electrocardiogram (ECG) Signals Related to Hypoglycaemia,‖ Conference on computers in Cardiology, pp. 537-540, IEEE, 2003.

[12] A. B. Ramli, and P. A. Ahmad, ―Correlation analysis for abnormal ECG signal features extraction,‖ 4th National Conference on Telecommunication Technology, 2003. NCTT 2003 Proceedings, pp. 232-237, 2003.

[13] Mazhar B. Tayel, and Mohamed E. El Bouridy, ―ECG Images Classification using Artificial Neural Network Based on Several Feature Extraction Methods,‖ IEEE, pp113-115, 2008.

[14] P. Tadejko, and W. Rakowski, ―Mathematical Morphology Based ECG Feature Extraction for the Purpose of Heartbeat Classification,‖ 6th International

Conference on Computer Information Systems and Industrial Management Applications, CISIM '07, pp. 322-327, 2007.

[15] Alan Jovic, and Nikola Bogunovic, ―Feature Extraction for ECG Time- Series Mining based on Chaos Theory,‖ Proceedings of 29th International Conference on Information Technology Interfaces, 2007.

[16] Ubeyli, and Elif Derya, ―Feature extraction for analysis of ECG signals,‖ Engineering in Medicine and Biology Society, EMBS 2008. 30th Annual International Conference of the IEEE, pp. 1080-1083, 2008.

[17] S. Z. Fatemian, and D. Hatzinakos, ―A new ECG feature extractor for biometric recognition,‖16th International Conference on Digital Signal Processing, pp. 1-6, 2009.

[18] Pedro R. Gomes, Filomena O. Soares, J. H. Correia, C. S. Lima ―ECG Data-Acquisition and Classification System by Using Wavelet-Domain Hidden Markov Models‖ 32nd Annual International Conference of the IEEE EMBS Buenos Aires, Argentina, August 31 - September 4, 2010.

[19] S. M. Jadhav, Dr. S. L. Nalbalwar, Dr. Ashok A. Ghatol, ―modular neural network based arrhythmia classification system using ecg signal data‖, in International Journal of Information Technology and Knowledge Management January-June 2011.